\documentclass{article}
\usepackage{spconf,amsmath,graphicx}
\usepackage{amssymb}

\usepackage{xcolor}

\usepackage{enumitem}

\newcommand{\ieno}{\textit{i}.\textit{e}.}
\title{A one-shot texture-perceiving generative Adversarial network for unsupervised surface inspection}
\name{Lingyun Gu$^{\dagger}$ \qquad Lin Zhang$^{\star}$ \qquad Zhaokui Wang$^{\dagger}$}
  
\address{ Tsinghua University, Beijing, China$^{\dagger}$\\
          University of Cincinnati, Cincinnati, Ohio, USA$^{\star}$ \\}
\begin{document}
% \ninept
%
\maketitle
\begin{abstract}
Visual surface inspection is a challenging task owing to the highly diverse appearance of target surfaces and defective regions. Previous attempts heavily rely on vast quantities of training examples with manual annotation. However, in some practical cases, it is difficult to obtain a large number of samples for inspection. To combat it, we propose a hierarchical texture-perceiving generative adversarial network (HTP-GAN) that is learned from the one-shot normal image in an unsupervised scheme. Specifically, the HTP-GAN contains a pyramid of convolutional GANs that can capture the global structure and fine-grained representation of an image simultaneously. This innovation helps distinguishing defective surface regions from normal ones.
%this novel HTP-GAN model contains a pyramid of convolutional GANs, which utilize a single image to simultaneously perceive its global structure and fine-grained representation. This indicates that a well-trained discriminator within HTP-GAN is sensitive to the defective regions.
In addition, in the discriminator, a texture-perceiving module is devised to capture the spatially invariant representation of normal image via directional convolutions, making it more sensitive to defective areas. Experiments on a variety of datasets consistently demonstrate the effectiveness of our method.
% Visual surface inspection is a challenging task owing to the highly diverse appearance of target surfaces and defective regions. Previous attempts heavily rely on vast quantities of training examples with manual annotation. However, in some practical cases, it is difficult to obtain large amount of samples for inspection.
% To combat it, we propose a hierarchical texture-perceiving generative adversarial network (HTP-GAN) that is learned from the a one-shot normal image in an unsupervised scheme. 
% Specifically, this novel HTP-GAN model contains a pyramid of convolutional GANs, which utilize a single image to simultaneously perceive its global structure and fine-grained representation. This indicates that a well-trained discriminator within HTP-GAN is sensitive to the defective regions. In addition, in the discriminator, texture-perceiving module is devised to capture the spatially invariant representation of normal image via directional convolutions, making it more sensitive to defective areas. Experiments on a variety of datasets consistently demonstrate the effectiveness of our method.
\end{abstract}
\section{Introduction}
\label{sec:intro}
Due to the rapid development of deep neural networks~\cite{2019SinGAN,20hierarchicalzheng20,zheng2021group,zheng2020exploiting}, visual surface inspection~\cite{zhai2018,1990Hierarchical} has attracted increasing attention as an important technology in many intelligent industrial applications.
Visual surface inspection aims to detect the abnormal regions on the surface of material using visual images.
It is a challenging task owing to various image noises, texture variations of the target surface, and highly diversified appearance of abnormal regions.  

Typical visual inspection approaches can be categorized into two main groups: traditional methods~\cite{1990Hierarchical,2007Verifying} and learning-based methods~\cite{zhai2018,2017Region,NEURIPS2019_f0031c7a,2017Fully}. Traditional methods adopt handcrafted features to perform surface inspection, which cannot be well generalized to new scenarios. Learning-based approaches achieve significant performance when equipped with large amounts of manually annotated training data. However, in some practical cases, such as surface inspection under planes ~\cite{jovanvcevic20173d}, only a small number of normal samples, or even a single sample, are available. Therefore, how to design the one-shot surface inspection method is very important for practical scenarios.
%\tcr{how to achieve unsupervised visual surface detection with as few images as possible is crucial.}
%However, the performance of the learning-based approaches usually degrades drastically when only few-shot samples or even one-shot sample are available. In some practical cases, such as surface inspection under planes or rockets, it is impractical to collect and annotate a tremendous amounts of defective examples. Therefore, how to design one-shot surface inspection methods is very important for practical scenarios where there is a lack of training data.

We present the problem of one-shot unsupervised surface inspection: given an example of a normal image as training data, all defective regions should be detected and segmented for arbitrary images with the same texture category as the normal image. The challenge lies in 1) how to design an adaptable perceiving model that is prone to handle the texture variations; 2) how to perform a more generalized surface inspection model in the one-shot way.

To deal with the challenge, we propose a novel hierarchical texture-perceiving generative adversarial network (HTP-GAN) that is learned from a one-shot normal image in an unsupervised scheme. 
Specifically, HTP-GAN model contains a pyramid of convolutional GANs which utilize a single image to simultaneously extract the global and fine-grained representation of the image. This guides the model to indeed learn the normal texture representation of the category of the corresponding image and can distinguish the defective surface regions.
In addition, in the discriminator of HTP-GAN, texture-perceiving module is devised to capture the spatially invariant representation of normal image via directional convolutions, making it more sensitive to defective areas. Experiments on a variety of datasets consistently demonstrate the effectiveness of our method.

\begin{figure*}[t!]
		\centering
		\includegraphics[width=0.8\linewidth]{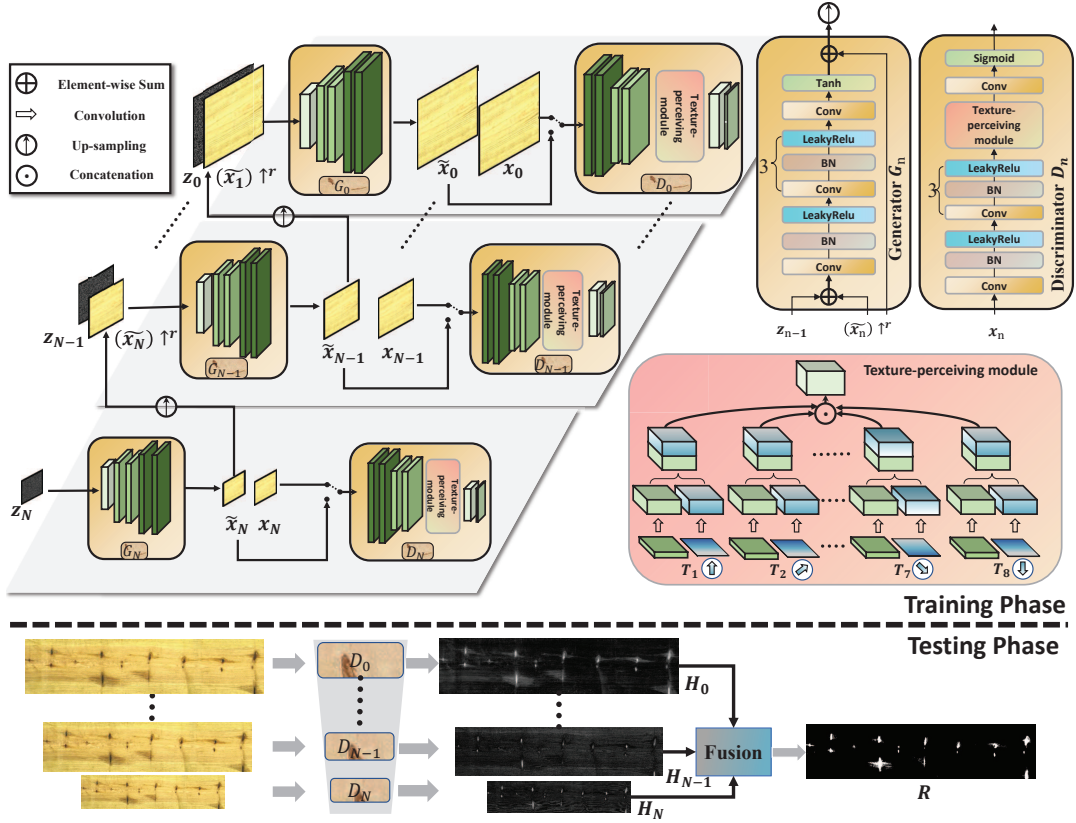}
		\caption{Overview of the proposed hierarchical texture-perceiving generative adversarial network (HTP-GAN). In the training phase, the HTP-GAN is devised to capture the spatially invariant representation of a single normal image via directional convolutions at multiple scales.
		In the testing phase, the well-trained discriminators can indeed memorize the latent distribution properties of normal texture representation and distinguish the defective surface regions.}
		\label{fig:method}
\end{figure*}

%an adaptable and texture-perceiving model is demanded to learn both global structure and fine details of the abnormal image example. In addition, global context is also an important cue as the surface texture is usually not orderless. So the subtle balance between local details and global context should be concerned. In this paper, we propose a hierarchical  texture-perceiving generative adversarial network (HTP-GAN) for one-shot unsupervised visual surface inspection, in which just single label-free samples is required. 

We summarize our main contributions as follows: 
	
\begin{itemize}[leftmargin=*,noitemsep,nolistsep]

\item  We introduce a novel one-shot surface inspection network with a pyramid of convolutional GANs, achieving unsupervised surface inspection with a single example of normal image.

\item A texture-perceiving module is devised to capture the spatially invariant representation of normal image via directional convolutions, making it more sensitive to defective areas.
%We propose a texture-perceiving module to learn the global context of normal surface in the latent feature space, resulting in robust spatially invariant representation.

\item Experimental results achieve the state-of-the-art performance on two public datasets, which demonstrate the effectiveness of the proposed approaches.

\end{itemize}

\section{Hierarchical Texture-perceiving Generative Adversarial Network}
\label{sec:pagestyle}

% \subsection{Problem setup}
Our goal is to detect the abnormal regions on the surface of material from a single image. When large amounts of training data are available, a well-trained generative adversarial network (GAN) can easily learn a representation of the distribution of the target samples via a generator $G$ and a discriminator $D$. But in the case of the limited number of samples, conventional GAN is hard to learn such a representation due to insufficient training data ~\cite{zhao2020differentiable}. In the context of learning from a single image, inspired by the SinGAN~\cite{2019SinGAN}, we adopt the downsampling strategy to generate different scales of images from a single image as training samples. Then the network is able to learn internal distributions at different scales.
At the same time, a pyramid of convolutional GANs is utilized to simultaneously capture the statistics of complex image structures and perceive the fine-grained representation of the normal image. To our best knowledge, this is the first time that the SinGAN being applied in the one-shot surface inspection task. 

% this is the first time that the multi-scale GANs being applied in the surface inspection task.

%While similar multi-scale architectures have been explored in the image generation task~\cite{2019SinGAN}, we are the first explore it in one-shot surface inspection task.

% The generator $G$ learns a distribution of normal surface data $x$ by a mapping $G(z)$ of samples $z$. The discriminator $D$ learns to distinguish generated samples $G(z)$ from real observation $x$. $G$ and $D$ are alternately optimized through the following function:

% At the test stage, we input the abnormal image into the discriminator $D$, 
% $G$ is trained to generate fake images analogous to the normal surface samples while $D$ is simultaneously trained to estimate the probability that a sample is from the training normal surface data. We except that 

\subsection{Hierarchical Fully Convolution Architecture}
\label{ssec:subhead}
In this section, we will detail the overall architecture of the proposed HTP-GAN. As shown in Fig.~\ref{fig:method}, the network is based on SinGAN~\cite{2019SinGAN} to achieve the one-shot learning task.
We change the traditional GAN from two perspectives: 1) we resize the input image to generate the multi-scale images. Specifically, given an image of normal surface $x_0$, we adopt the down-sampled strategy to generate an image pyramid of $x: \{x_0, \ldots , x_N\}$. Multi-scale inputs contain more fine details and texture information.
2) a pyramid of convolutional GANs are utilized to simultaneously capture the characteristics of complex image structures and perceive fine-grained representation of the training image. So we need to design a pyramid of generators $\{G_0, \ldots, G_N \}$ and discriminators $\{D_0, \ldots, D_N \}$ to handle the multi-scale inputs. 

At the training phase, the proposed HTP-GAN is a multi-stage training process from coarse-grain to fine-grain. Firstly, a noise map is injected into the generator $G_N$ at the coarsest scale to generate the $\tilde{x}_{N}$. Then a combination between the generated image and the noise map sequentially passes through all other generators up to the finest scale:

\begin{equation}
\resizebox{.9\hsize}{!}{$\tilde{x}_{n}=G_{n}\left(z_{n},\left(\tilde{x}_{n+1}\right) \uparrow^{r}\right)=\left(\tilde{x}_{n+1}\right) \uparrow^{r}+\psi_{n}\left(z_{n}+\left(\tilde{x}_{n+1}\right) \uparrow^{r}\right)$},
\end{equation}
where $\psi_{n}$ is a fully convolutional net with 5 conv-blocks and $\uparrow^{r}$ denote to the up-sampling operation. As shown in Fig.~\ref{fig:method}, the first 4 conv-blocks consist of a $3\times3$ convolutional layer following with a BatchNorm and a LeakyReLU. The last conv-block consist of a $3\times3$ convolutional layer and a tanh activation layer. In addition, the $\psi_{n}$ is able to learn the residual feature of a generated image at a finer scale.

\subsection{Textual-perceiving Discriminator}
%\tcr{To capture the patch distribution in the corresponding image $x_n$, the discriminator is adopted to perform adversarial learning with the generator. It is worth mentioning that conventional discriminator dose not propose a corresponding module for texture perception.}
A textual-perceiving module in the discriminator is devised to capture the spatially invariant representation of a normal image via directional convolutions, making it more sensitive to defective areas. Specifically, the textual-perceiving discriminator consists of the several conv-blocks $f_n$, the textual-perceiving module~\cite{zhu2019one} ($c_n$, $g_n$) and a sigmoid activation layer. As shown in Fig.~\ref{fig:method},  the several conv-blocks $f_n$ are used to extract preliminary features $F_n$ from the generated image. Then, the $c_n$ and $g_n$ are designed to capture the spatially invariant representation of normal image via eight directional texture features under the guidance of corresponding directional feature map $T_i$. Specifically, 

\begin{equation}
P_{n}=c_{n}\left(\operatorname{cat}\left(f_{n}(\tilde{x}_{n}), T_{i}\right)\right)(i=1,2 \cdots 8),
\end{equation}
where $c_n$ denotes several convolutions which contains a $3\times3$ convolution layer with a BatchNorm and a ReLU activation layer, $cat$ denotes the concatenation operation. In our design, total eight directions are adopted, including top, bottom, left, right, top left, bottom left, top right, and bottom right. Each directional map $T_i$ is a generated trend square matrix that decreases from 1 to 0 in a certain direction. Finally, the output features of different branches are concatenated to form the whole spatial invariant feature $M_{n}$, that is,

\begin{equation}
M_{n}=Sigmoid(g_{n}\left(\operatorname{cat}\left(P_{1}, P_{2} \cdots P_{8}\right)\right)),
\end{equation}

where $g_n$ represents a set of standard convolution block, $M_{n} \in \mathbb{R}^{1 \times H \times W}$ is the distinguish map that is used to calculate the loss of discriminator. Since the proposed directional convolution unit is sensitive to local variations of the image along each direction, it can make the network well adaptable to spatial distortions and scale variations. Intuitively, the discriminator of a well-trained GAN memories the patch distribution of a normal image. So it should be insensitive to the normal regions but varies drastically in the abnormal regions.

\subsection{Hierarchical Fusion for Surface Inspection}

In this section, we introduce a hierarchical fusion strategy to fuse the multi-scale distinguish maps for producing the final segmented result. 

Inspired by~\cite{zhai2018}, the information entropy is a suitable metric to represent the output of discriminator for abnormal region segmentation. Thus, the information entropy of multi-scale distinguish maps are expressed as following:
\begin{equation}
H_{n} = M_{n}*\log M_{n},
\end{equation}

$H = \{H_{n}\}|_{n=0}^{N}$ are a set of information entropy of multi-scale distinguish maps and we refer to $H$ as the coarse inspection map set, which reveals the coarse abnormal regions of the input image.

The GANs have small receptive fields and limited capacity, preventing them from representing the single image. To capture global textual structure and fine-grained texture information, we fuse the multi-scale coarse inspection maps together to better distinguish the abnormal regions. Thus, the fusion process expresses as following:
\begin{equation}
R =  \sum_{n=0}^{N} \alpha_n * H_{n},
\end{equation}
where $R$ denotes the final fusion and $\alpha_n = \frac{1}{N+1}$ are weighting factors.

\section{experiments}
\label{sec:typestyle}

To evaluate the effectiveness of HTP-GAN in one-shot surface inspection task, we design extensive experiments based on WOOD Defect Database (WOOD)~\cite{Olli2003Wood} and Road Crack Database (CRACK)~\cite{7025160}. Furthermore, we compare our method with four representative methods as following: (1) Unsupervised visual surface inspection method proposed in~\cite{zhai2018} (ICASSP 2018); (2) Surface defect detection method based on positive samples and artificial defects~\cite{2018A} (Prical 2018); (3) Automated surface inspection method proposed in ~\cite{2019Segmentation} (JOIM 2019); (4) Semantic image segmentation method proposed in \cite{chen2018encoder} (ECCV 2018). 

\subsection{ Datasets and Evaluation Metrics}
{\bf Dataset and Setting}: 
WOOD Defect Database (WOOD) and Road Crack Database (CRACK) are two surface inspection datasets which are annotated with segmentation labels of abnormal regions~\cite{zhai2018}.
For our HTP-GAN, a single image with normal surface is adopted as the training data, which is the embodiment of the one-shot surface inspection. To compare our method with unsupervised methods:~\cite{zhai2018} and~\cite{2018A}, we adopt the same image with normal surface to train the models. 
For supervised methods:~\cite{2019Segmentation} and~\cite{chen2018encoder}, several abnormal samples and normal samples are used to train the networks. All images are resized to 256 $\times$ 256 size.

\noindent {\bf Evaluation Metrics}: We adopt two evaluation metrics to compare and analyze experimental results: Intersection-over-union (IOU) and Pixel Accuracy (pixel acc).

\vspace{-10pt}

\subsection{Comparison with the State-of-the-art Approaches}

\begin{figure}[t!]
		\centering
		\includegraphics[width=1.0\linewidth]{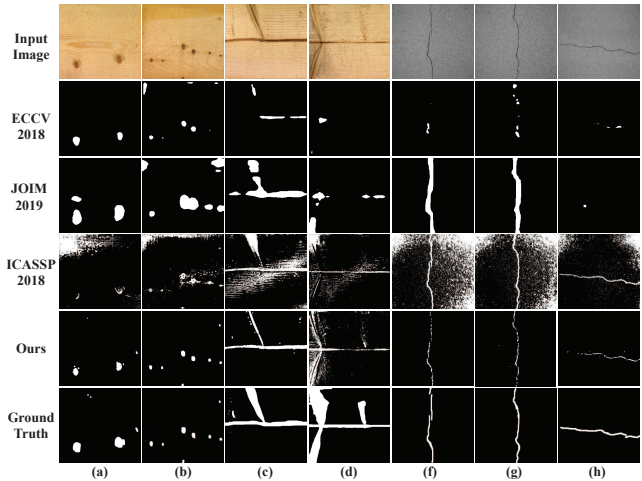}
		\caption{Visualization comparison with the state-of-the-art methods for the surface inspection task.
		}
		\label{fig:result}
\vspace{-10pt}
\end{figure}

\begin{table}[]
\footnotesize
\centering
\caption{Quantitative comparisons. (IoU(\%) / pixel acc(\%))}
\begin{tabular}{c|c|c|c|c}
\hline
      & \cite{zhai2018} & \cite{chen2018encoder} & \cite{2019Segmentation} & Ours \\ \hline
 WOOD  & 44.83/81.96  & 47.23/94.45 & 47.37/93.82 & \bf59.83/96.54\\ 
 CRACK &  31.06/58.21   & 48.68/90.46  & 33.02/61.56  & \bf56.97/96.32 \\ \hline
\end{tabular}
\label{TAB:1}
\end{table}

% \begin{table}[]
% \footnotesize
% \centering
% \caption{Quantitative comparisons. (IoU(\%) / pixel acc(\%))}
% \begin{tabular}{c|c|c|c|c}
% \hline
%       & \cite{zhai2018} & \cite{chen2018encoder} & \cite{2019Segmentation} & Ours \\ \hline
%  WOOD  & 44.83/81.96  & 47.23/94.45 & 47.37/93.82 & \textbf{59.83}/\textbf{96.54}\\ 
%  CRACK &  31.06/58.21   & 48.68/90.46  & 33.02/61.56  & \textbf{56.97}/\textbf{96.32}  \\ \hline
% \end{tabular}
% \label{TAB:1}
% \end{table}

Tab.~\ref{TAB:1} shows the performance comparison of our HTP-GAN against the other four methods in terms of IoU and pixel acc on the WOOD and CRACK dataset. As we can see in Tab.~\ref{TAB:1}, our method achieved higher IoU and pixel acc than the four methods.
  
In the one-shot setting, when the single normal image or an image pair (a normal image and an abnormal image) is given as training data, all four methods can not predict any abnormal regions or produce a noise map. This result demonstrates that these methods rely on a tremendous amount of training data and these models are easy to cause overfitting on one-shot surface inspection task. Thus, we try to add the training data on these methods in the supervised learning methods or adopt the fully convolution network to improve the unsupervised learning methods~\cite{2019SinGAN}.

Fig.~\ref{fig:result} shows the visualization of different methods for surface inspection. ~\cite{2019Segmentation} has a good performance until we adopt 10 normal-abnormal image pairs for training while ~\cite{chen2018encoder} performs well on 4 abnormal images. 
Although improved \cite{zhai2018} detects the more accurate abnormal surface regions than other methods, our proposed model has a better performance. More importantly, HTP-GAN only needs a normal image as training data and achieves the best performance.
%While the generic unsupervised approaches ICASSP 2018 and Prical 2018 need more than one training image, besides supervised approach ECCV 2018 and JOIM 2019 need 10 normal-abnormal image pairs for training process.
 
 %As shown in Fig. 5, our HTP-GAN model detected almost all of the detective surface with one-shot training.

\vspace{-10pt}

\subsection{Ablation Studies}

\begin{figure}[t!]
    \small
		\centering
		\includegraphics[width=0.8\linewidth]{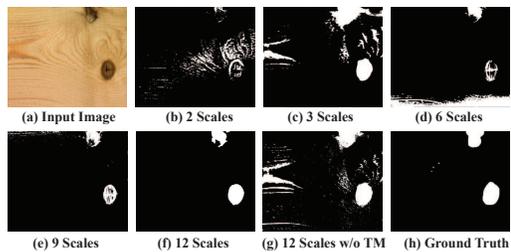}
		\caption{Visualization of influence of different scales and textual-perceiving module in our proposed HTP-GAN. ``TM'' refers to the textual-perceiving module.%We show the effect of starting our HTP-GAN from a given level $n$. For our full scheme ($n = 12$), our model achieves the best performance.
		}
		\vspace{-10pt}
		\label{fig:ab}
\end{figure}

% \begin{figure}[h!]
%     \small
% 		\centering
% 		\includegraphics[width=0.8\linewidth]{abli1.eps}
% 		\caption{Influence of textual-perceiving module. ``TM'' refers to the textual-perceiving module.
% 		}
% 		\vspace{-10pt}
% 		\label{fig:ab1}
% \end{figure}

\noindent\textbf{Effectiveness of Scale Number.}  
The number of scales in HTP-GAN architecture has a strong influence on the results. A small number of scales only can capture the local textures, leading to poor segmentation results. As the number of scales increases, Fig.~\ref{fig:ab} (a)-(f) demonstrates that our proposed method manages to capture larger structures as well as the fine-grained representation of the image. It indicates that a strong representation can help the model to segment most of the anomalous regions very well.

\noindent\textbf{Effectiveness of Textual-perceiving Module.}
% In the discriminator of HTP-GAN, texture-perceiving module is devised to capture the spatially invariant representation of normal image via directional convolutions, making it more sensitive to defective areas.
As shown in Fig.~\ref{fig:ab} (g) and (f), by adding the textual-perceiving module, our method segments more abnormal regions and alleviates to split normal regions into defects. It demonstrates that the proposed textual-perceiving module can learn spatially invariant representation of normal images and be more sensitive to defective areas.
 
\noindent\textbf{Influence of Different Image Variations.}
Our proposed method can handle most variations (\ieno Translation, Mirror, Scaling). However, for rotation variation, when the training image contains mostly horizontal textures (such as the wood in Fig.~\ref{fig:ab}, a 45-degree change in rotation only brings a slight performance drop 3\% in term of IOU score, and no performance drop in term of pixel accuracy.
% Because our framework only uses a single image with a horizontal texture to train the model, which is hard to capture the characteristics of rotation invariance. 
% When introducing more rotation augmentation, our method is not sensitive to the rotation variation.
 
\section{Conclusion}
In this paper, we propose a hierarchical texture-perceiving generative adversarial network (HTP-GAN) to achieve the one-shot surface inspection task.
%s to perceive the global and fine-grained representation
By applying a pyramid of convolutional GANs, HTP-GAN can indeed learn the global and fine-grained representation of normal surface from a single image. Thus, it enables discriminators in well-trained HTP-GAN to exhibit more active responses for the defective regions. Furthermore, the texture-perceiving module is devised to capture the spatially invariant representation of normal, making the discriminator more sensitive to defective areas.
% The performance of our method dominates the comparison against existing advanced algorithms by means of intersection-over-union and pixel accuracy. 
More importantly, the training of the HTP-GAN only relies on a single image, which explores a more general and practical solution to the surface inspection task.
%Experimental results consistently demonstrate the effectiveness of our method.

% this novel HTP-GAN model utilize single normal image to simultaneously perceive the global structure and fine-grained representation of the normal training image via a pyramid of convolutional GANs. This indicates that a well-trained HTP-GAN can indeed learn a good representation of normal surface images and is sensitive to the defective regions. 
% In addition, in the discriminator of HTP-GAN, texture-perceiving module is devised to capture the spatially invariant representation of normal image via directional convolutions, making it more sensitive to defective areas. Experiments on a variety of datasets consistently demonstrate the effectiveness of our method.

% \section{Acknowledgements}
% \label{sec:pagestyle}
% This work was conducted within the Key Laboratory of Intelligent Infrared Perception, Chinese Academy of Sciences.

\section*{Acknowledgments}
This work was supported by the Key Laboratory of Intelligent Infrared Perception, Chinese Academy of Sciences.

\bibliographystyle{IEEEbib}
\bibliography{strings,refs}

\end{document}